\documentclass[10pt,conference]{IEEEtran}
\IEEEoverridecommandlockouts
\usepackage{cite}
\usepackage{amsmath,amssymb,amsfonts}
\usepackage{algorithmic}
\usepackage{graphicx}
\usepackage{textcomp}
\usepackage{xcolor}
\usepackage{subcaption}

\def\BibTeX{{\rm B\kern-.05em{\sc i\kern-.025em b}\kern-.08em
    T\kern-.1667em\lower.7ex\hbox{E}\kern-.125emX}}
\usepackage{geometry}
 \IEEEoverridecommandlockouts\IEEEpubid{\makebox[\columnwidth]{ 978-1-6654-3540-6/22~\copyright~2022 IEEE \hfill} \hspace{\columnsep}\makebox[\columnwidth]{ }}
\begin{document}

\title{Efficient Information Sharing in ICT Supply Chain Social Network via Table Structure Recognition \\
\thanks{This work was supported in part by Mathematics of Information Technology and Complex Systems (MITACS) Accelerate Program, Smart Computing for Inovation Program (SOSCIP) and Lytica Inc.}
}

\author{\IEEEauthorblockN{Bin Xiao, Yakup Akkaya, Murat Simsek, Burak Kantarci}
\IEEEauthorblockA{\textit{School of Electrical Engineering and Computer Science} \\
\textit{University of Ottawa}\\
Ottawa, Canada \\
\{bxiao103, yakka087, murat.simsek, burak.kantarci\}@uottawa.ca}
\and

\IEEEauthorblockN{Ala Abu Alkheir}
\IEEEauthorblockA{\textit{Directorate, Analytics} \\
\textit{Lytica Inc.}\\
Ottawa, Canada \\
ala\_abualkheir@lytica.com}
}

\maketitle

\begin{abstract}

The global Information and Communications Technology (ICT) supply chain is a complex network consisting of all types of participants. It is often formulated as a Social Network to discuss the supply chain network’s relations, properties, and development in supply chain management. Information sharing plays a crucial role in improving the efficiency of the supply chain, and datasheets are the most common data format to describe e-component commodities in the ICT supply chain because of human readability. However, with the surging number of electronic documents, it has been far beyond the capacity of human readers, and it is also challenging to process tabular data automatically because of the complex table structures and heterogeneous layouts. Table Structure Recognition (TSR) aims to represent tables with complex structures in a machine-interpretable format so that the tabular data can be processed automatically. In this paper, we formulate TSR as an object detection problem and propose to generate an intuitive representation of a complex table structure to enable structuring of the tabular data related to the commodities. To cope with border-less and small layouts, we propose a cost-sensitive loss function by considering the detection difficulty of each class. Besides, we propose a novel anchor generation method using the character of tables that columns in a table should share an identical height, and rows in a table should share the same width. We implement our proposed method based on Faster-RCNN and achieve 94.79\% on mean Average Precision (AP), and consistently improve more than 1.5\% AP for different benchmark models.

\end{abstract}


\section{Introduction}
\label{sec:introduction}
\begin{figure}
     \centering
     \begin{subfigure}[b]{\columnwidth}
         \centering
         \includegraphics[width=\columnwidth]{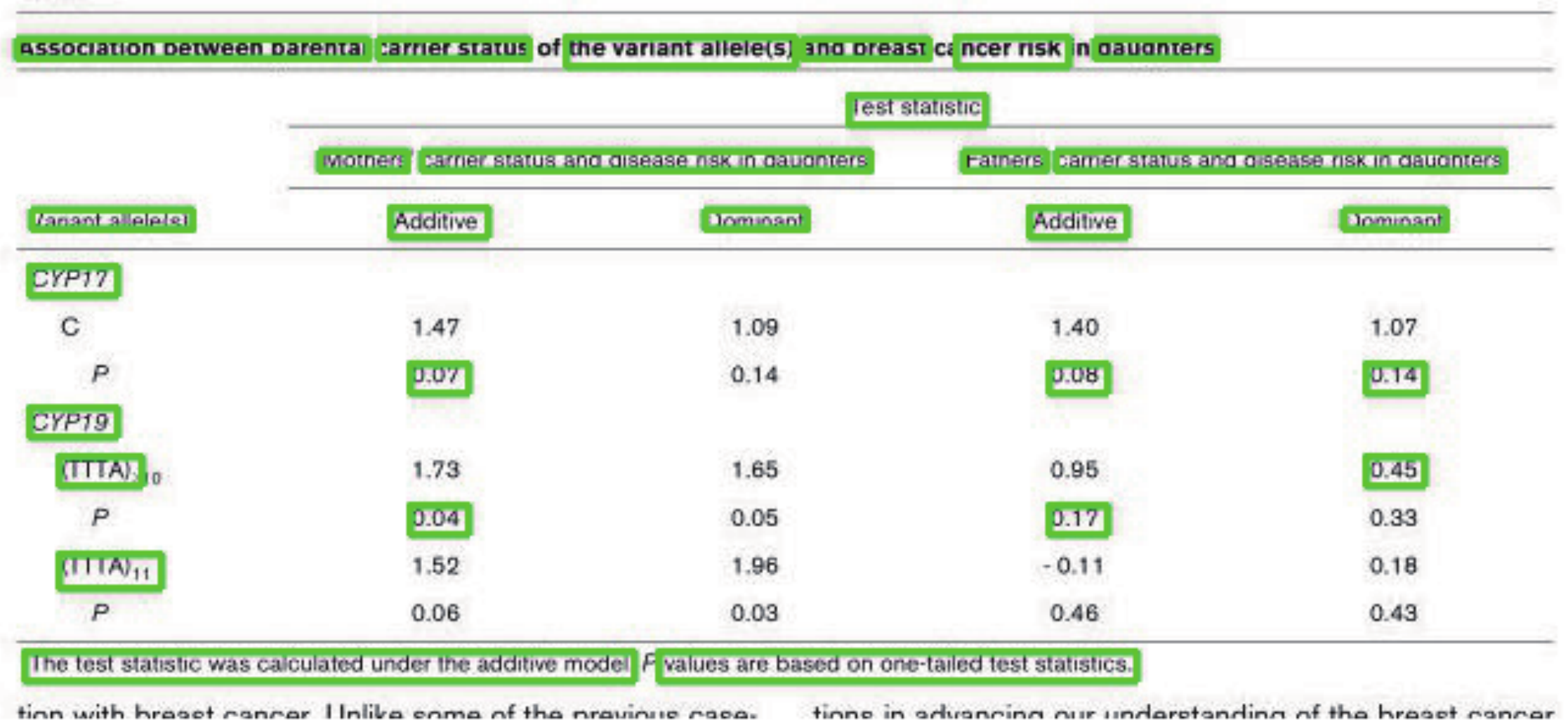}
         \caption{}
         \label{fig:easyocr_sample}
     \end{subfigure}
     \hfill
     \begin{subfigure}[b]{\columnwidth}
         \centering
         \includegraphics[width=\columnwidth]{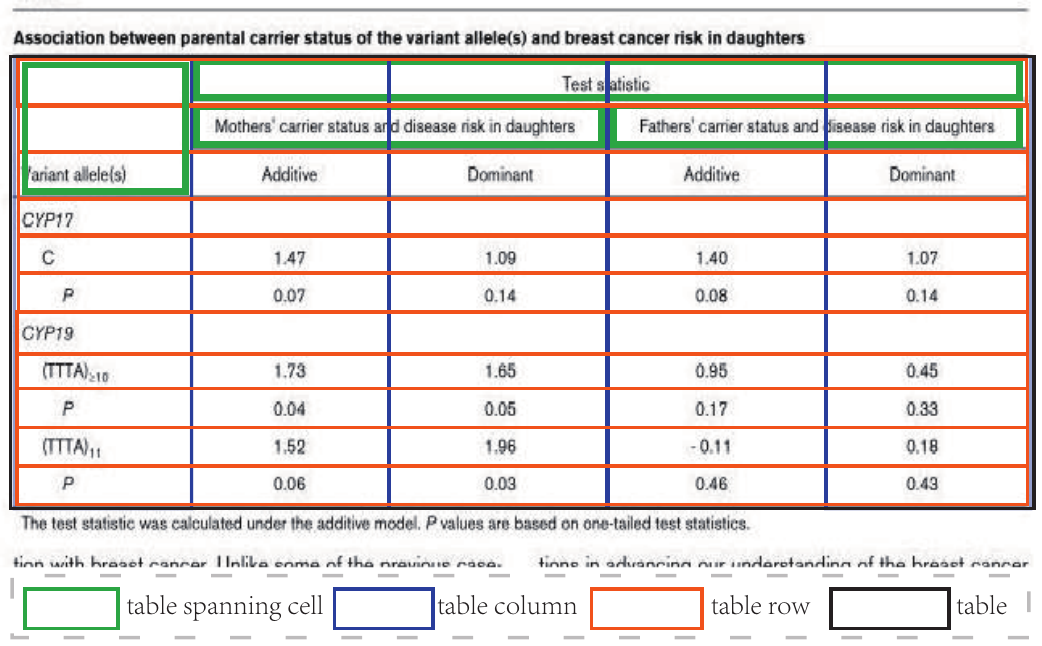}
         \caption{}
         \label{fig:pubtable1m_sample}
     \end{subfigure}
    \caption{Figure (a) shows a sample output of EasyOCR, in which all the detected areas all content-focused, some cells are recognized as multiple cells, and some cells are not detected. Figure (b) shows the defined four types of objects in this paper.}
    \label{fig:table_sample}
\end{figure}
The global Information and Communications Technology (ICT) supply chain is a complex network consisting of individuals, organizations, countries, suppliers, consumers, and other participants. This complex supply chain network is often formulated as a Social Network and analyzed by Social Network Analysis tools in the supply chain management and optimization~\cite{rodriguez2016social}. Information sharing plays a critical role in the ICT supply chain optimization~\cite{lotfi2013information}, and datasheets have been widely used to describe e-component commodities and to share information across different participants because of their user-friendliness to human readers. Besides, tables are also widely used in these datasheets to summarize and present important information.

Even though tables in datasheets are compact and convenient for human readers, it is impractical for human readers to deal with the vast number of electronic documents. Therefore, it becomes critical to detect and process the tables in datasheets automatically so as to extract, organize and share information in the supply chain network efficiently. This is still a challenging task because tables in datasheets are not structured, and the appearance of these tables varies. Figure~\ref{fig:table_sample} shows a table sample from PubTables1M~\cite{smock2021pubtables1m} dataset, which does not have explicit vertical lines among tables columns but contains spanning cells across multiple columns and rows, making it challenging to identify and locate the table cell precisely. Since datasheets cannot provide enough meta-data to extract and interpret tables and can be easily converted into document images, we focus on document images in this study, which is also the dominant problem assumption in recent studies. Most of the related studies~\cite{easyocr, du2021pp} use the margin among characters to determine whether they belong to an identical word or sentence. As a result, the detected results of these solutions suffer from recognizing the content in a single table cell as several cells, and the bounding boxes are content-focused, and cannot deal with empty cells, which can not reflect the complex table structures. Figure~\ref{fig:easyocr_sample} shows an example output of EasyOCR~\cite{easyocr}, in which all the bounding boxes are content-focused, some cells are broken into multiple different parts, and many cells are missed. In contrast, in this paper, following the design of the study in~\cite{smock2021pubtables1m}, we locate the table rows and columns instead of table cells and use the intersection of rows and columns to infer the location of table cells. As shown in Figure~\ref{fig:pubtable1m_sample}, table columns and table rows can help identify the contents in an identical column and row, respectively, and the intersection area of table columns and table rows can be inferred as table cells. Besides, to deal with complex table structures, table spanning cell is defined so that some cell areas can be merged to recover the original table structure. 

\begin{figure}[htp]
\begin{center}
  \includegraphics[width=0.9\columnwidth]{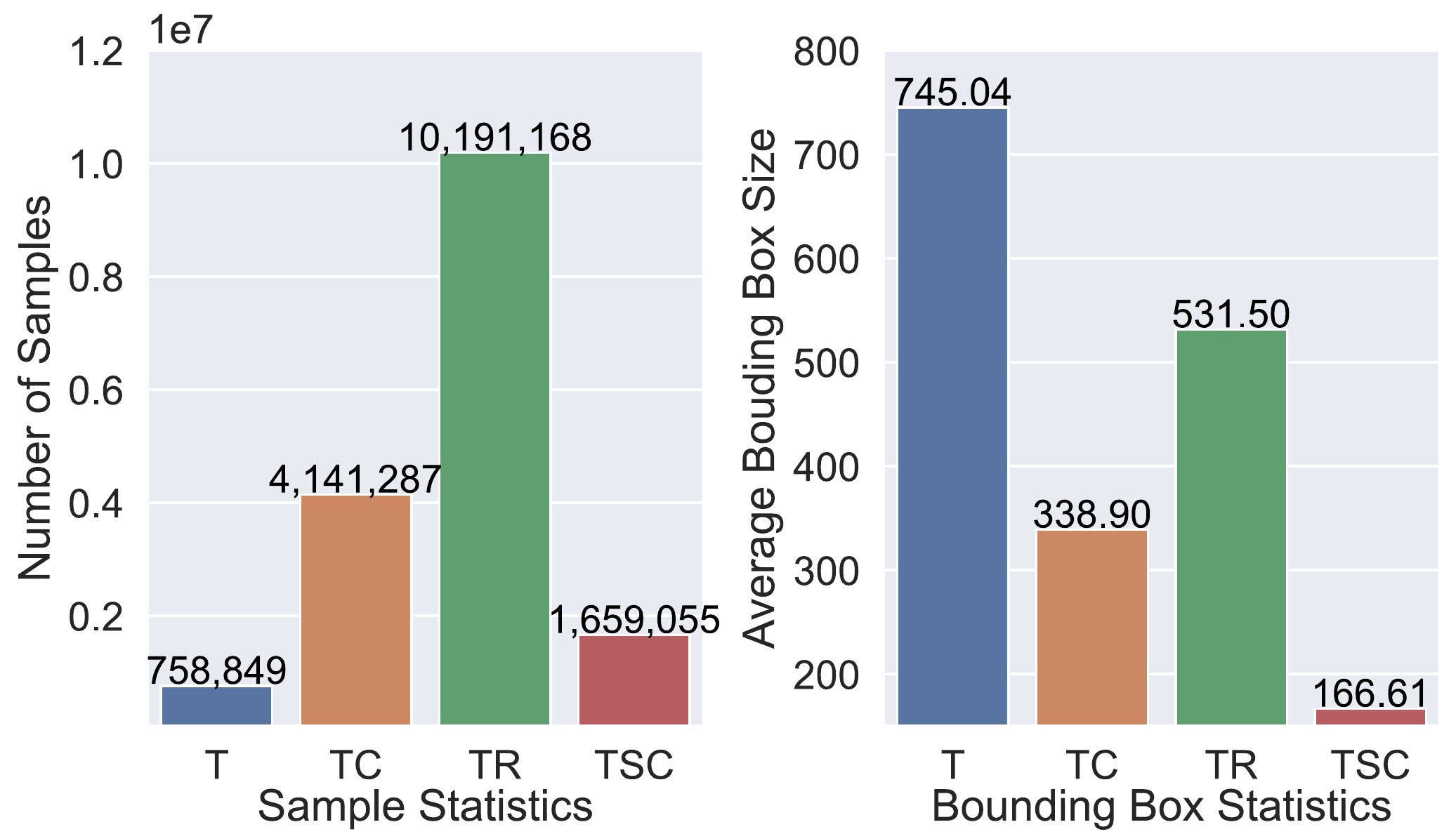}
  \caption{Statistics of samples and bounding boxes. Notably, T, TC, TR and TSC are short for table,  table column, table row and table spanning cell respectively, and the size of a bounding box is defined as the sum of its height and width.} 
  \label{fig:dataset_statis}
\end{center}
\end{figure}

On the other hand, object detection models, such as Faster-RCNN~\cite{ren2015faster}, Mask-RCNN~\cite{he2017mask}, have also been applied to the Table Detection (TD) and table cell detection problems in many studies. Typically, these two-stage object detection methods need to generate a set of region proposals by the Region Proposal Network (RPN) and then classify and reduce the number of region proposals by combining the classification and regression tasks. Since objects in an image can have different sizes, one common weakness of such methods is that small objects usually cannot be detected as accurately as large objects. Besides, the number of samples in each category is usually imbalanced. Figure~\ref{fig:dataset_statis} shows the statistics of the PubTable1M~\cite{smock2021pubtables1m} dataset, which is a large scale dataset used in Section~\ref{sec:experiments}. The distribution of each class's samples and the average size of each class's bounding boxes are extremely imbalanced. Therefore, in this study, we define the detection difficulty of each class by considering the sample amount and the average bounding box size of each class and propose a detection difficulty based cost-sensitive loss to balance each class to be detected in both the classification task and the regression task in the two-stage object detection models. Moreover, it is straightforward to observe that all columns in a table share an identical height, and all rows in a table have an identical width in our problem definition. Based on this observation, we propose a new anchor generation method that generates a series of fixed height bounding boxes for columns and a series of fixed width boxes for rows and sampling the region proposals with an imbalanced policy giving small objects higher priorities.

At last, even though there have been studies discussing the TSR problem, most of them only use some small-scale datasets. PubTable1M~\cite{smock2021pubtables1m} is a large dataset containing 758K tables with the annotation of columns and rows. In this study, we use this large dataset to conduct experiments and further analyze the impact of different number of training samples. To sum up, the contribution of this work can be three folds: 
\begin{enumerate}

\item Following the definition of study~\cite{smock2021pubtables1m}, we define four types of objects in a table and propose a top-down approach to represent the complex table structure, make the unstructured tables structured and machine-interpretable, and conduct experiments in a large scale, further analysis the impact of different training samples. We hypothesize that this proposed method will improve the information sharing efficiency in the context of ICT supply chain Social Network of commodities.

\item We propose a detection difficulty based cost-sensitive loss function that can guide the model to learn more features from the classes which are hard to be predicted and located. 
\item We propose a new anchor generation method based on the observation that all the columns in a table have a fixed height and all the rows in a table share a fixed width in our problem definition. Our experimental results demonstrate that our proposed method can consistently improve the benchmark models by around 1.5\% Average Precision (AP), and our trained Faster-RCNN based model can achieve 94.79\% AP.
\end{enumerate}
The rest of this paper is organized as follows: Section~\ref{sec:related_work} discusses related studies, including studies in TD and TSR problems. Section~\ref{sec:proposed_method} presents the problem definition, and the proposed method. Section~\ref{sec:experiments} shows the experimental results and discusses the design aspects of the proposed method. At last, we draw our conclusion and possible directions in section~\ref{sec:conclusion}.

\section{Related work}
\label{sec:related_work}

As discussed in section~\ref{sec:introduction}, the global ICT supply chain network can be formulated as a complex Social Network, and tabular data has been the de facto standard for information sharing among the participants in the ICT supply chain community network. In this section, we discuss related studies on the TSR problem, which can be adopted to process tabular data in ICT supply chain, and some typical solutions to the data imbalance problem in the context of TSR.
\subsection{Table Detection and Table Structure Recognition}
TD and TSR problems have been widely discussed recently. Typically, TD problem is defined as a objection detection problem, and two-stage object detection models, such as Faster-RCNN~\cite{ren2015faster}, Mask-RCNN~\cite{he2017mask} and their variations, are more popular than one-stage methods to be adopted in this problem because two-stage models usually can achieve higher performance. Many design aspects, including backbone network, model architecture, and transfer learning method, augmentation methods can influence the performance of two-stage object detection models. CascadeTabNet~\cite{prasad2020cascadetabnet} is a variation of Mask-RNN, which uses HRNetV2p~\cite{wang2020deep} as the backbone and proposes an iterative transferring method in order to make the model can be adapted to small datasets. TableDet~\cite{fernandes2021tabledet} uses a proposed augmentation method named Table Aware Cutout (TAC), which is implemented by cutting out some tables in the document images randomly. The TAC method is applied to popular two-stage detection models and achieved promising results.

Table Structure Recognition (TSR) problem aims at representing the complex table structure with a machine-interpretable format and converting the unstructured tabular data into a structured format. In general, popular methods for the TSR problem can be roughly categorized into two groups: top-down approaches and bottom-up approaches. Typically, top-down approaches define the TSR problem by detecting the table columns and table rows directly. DeepDeSRT~\cite{schreiber2017deepdesrt} is a typical top-down approach that uses a segmentation method to segment table columns and table rows directly, which can be implemented by Fully Convolution Networks. Meanwhile, other studies~\cite{fernandes2021tabledet, smock2021pubtables1m } also utilized object detection method to detect table columns and table rows. Study~\cite{smock2021pubtables1m} evaluates the two-stage object detection model and transformer based detection model and sets up new benchmarks on a new dataset. In contrast, bottom-up approaches~\cite{xiao2022table, chi2019complicated} use table cells as the basic units in a table and use a graph to represent the complex structure. More specifically, in bottom-up approaches, table cells are represented by the graph nodes, and three types of cell associations including "No Connection", "Horizontal Connection" and "Vertical Connection" are defined to build the relations among cells, which can be represented by the edges in a graph. GraphTSR~\cite{chi2019complicated} is a typical bottom-up approach that proposes a graph attention model to build the relation among table cells with the assumption that all the cells' bounding boxes are known.  

\subsection{Data Imbalance and Hard example mining}
Dataset imbalance is a common problem when applying machine learning algorithms in many scenarios. Cost-sensitive learning and hard example mining methods are widely used to alleviate the dataset imbalance problem. Study~\cite{khan2017cost} propose a deep model cost-sensitive model that can learn robust features from both minority and majority samples by training the class-dependent cost and model parameters jointly. In object detection problems, the imbalance problem can also have adverse effects on the model performance. The hard example sampling method is a common method when training RPN network to generate balanced positive and negative samples considering the difficulty of samples, and the detection difficulty is often modeled by the confidence score or the IoU score. Study~\cite{shrivastava2016training} proposes an Online Hard Example Mining (OHEM) considering the loss values of positive and negative samplings during training. Libra R-CNN~\cite{pang2019libra} discusses the imbalance problem at the sample level, feature level, and objective level and proposes using balance losses using the IoU as the measurement of sample hardness. Study~\cite{oksuz2020imbalance} categorizes the imbalance problem into four groups, including class imbalance, scale imbalance, spatial imbalance, and objective imbalance, and further discusses various solutions to these groups of imbalance problems in the object detectors. 

\section{Proposed Method}
\label{sec:proposed_method}
In the global ICT supply chain network, crucial information, especially for e-component commodities, is often summarized and presented by tables in the datasheets. However, these tabular data is unstructured and not machine-readable, and the volume of data has been far beyond the capacity of human readers. In this section, we present the formal problem definition and our proposed method to transform these unstructured tabular into structured, machine readable format by solving the TSR problem to form ICT supplier communities from unstructured data with minimum human intervention.

\subsection{Problem Definition}
\label{sec:problem_definition}
In this study, we formulate the TSR problem as an object detection problem. Given a training set $\mathbf{D}_{train}=\{x_i,y_i^m,b_i^m\}_{i=1}^N$, where $N$ is the number of samples in the training set, $x_i$ is the $i$th image sample, $b_i^m$ is the $m$th bounding box of the $i$th image and $y_i^m$ is the corresponding object class of $b_i^m$. We define four classes of objects in our TSR problem, namely table, table column, table row, table spanning cell, each of which can be denoted by a number, namely, $y_i^m \in \{0,1,2,3\}$. The target of the problem is to train a model that can predict the defined objects' bounding boxes in a image and their classes, which means that $P_\theta\{b_k^m = \hat{b}_k^m, y_k^m = \hat{y}_k^m \rvert x_k\}$. Notably, a bounding box can be denoted by a coordinate, a height, and a width, meaning that $b_i^m=\{x_i^m, y_i^m, width_i^m, height_i^m\}$. It is a regression problem to predict the correct bounding boxes, and it is a classification problem to predict the type of the object in each bounding box. 

\begin{figure*}[htp]
\begin{center}
  \includegraphics[width=0.7\textwidth]{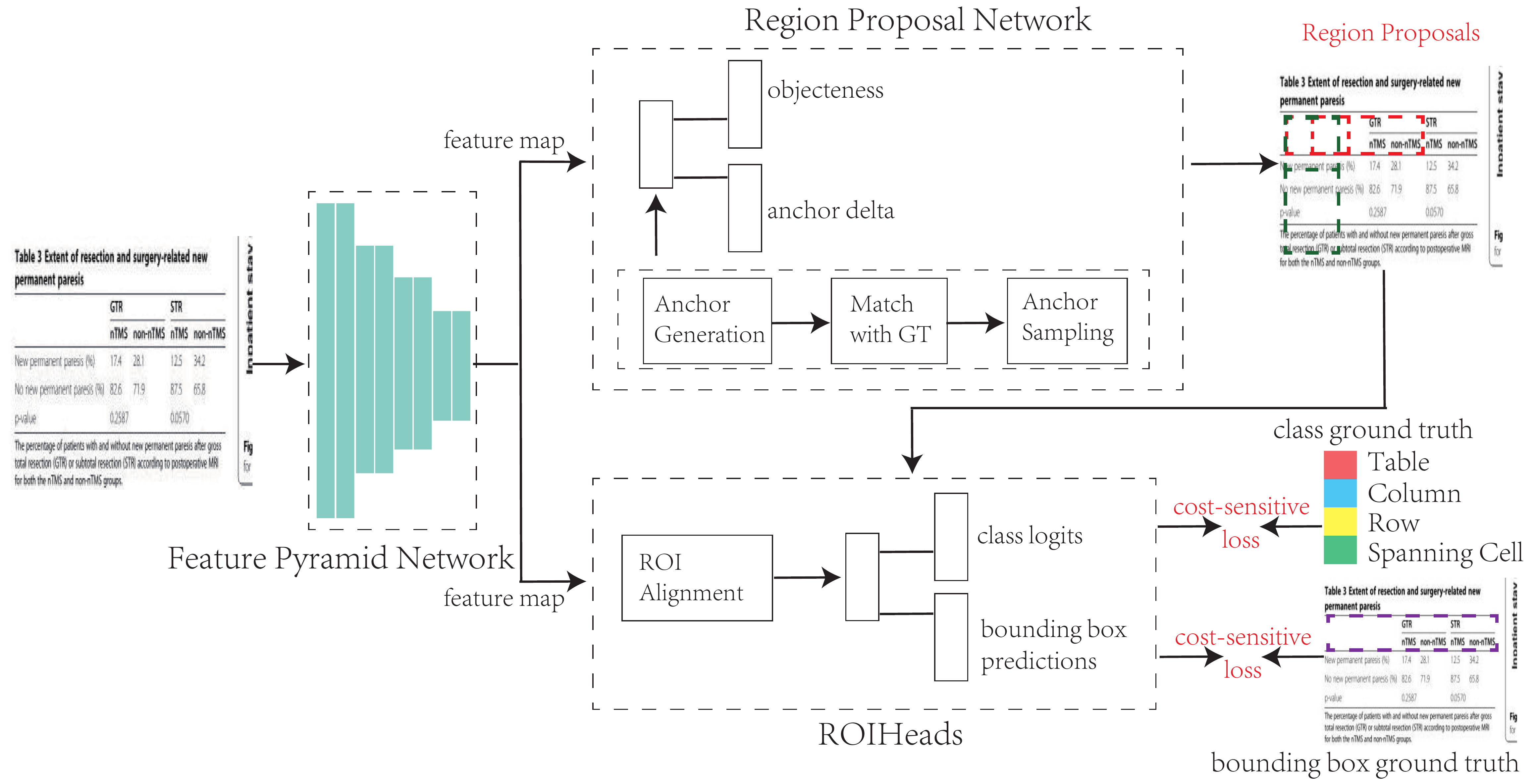}
  \caption{Overall architecture of generalized Faster R-CNN.}
  \label{fig:rcnn_architecture}
\end{center}
\end{figure*}

\subsection{Structure Aware Anchor Generation}
\label{sec:structure_aware_anchor_generation}
\begin{figure}[htp]
\begin{center}
  \includegraphics[width=0.7\columnwidth]{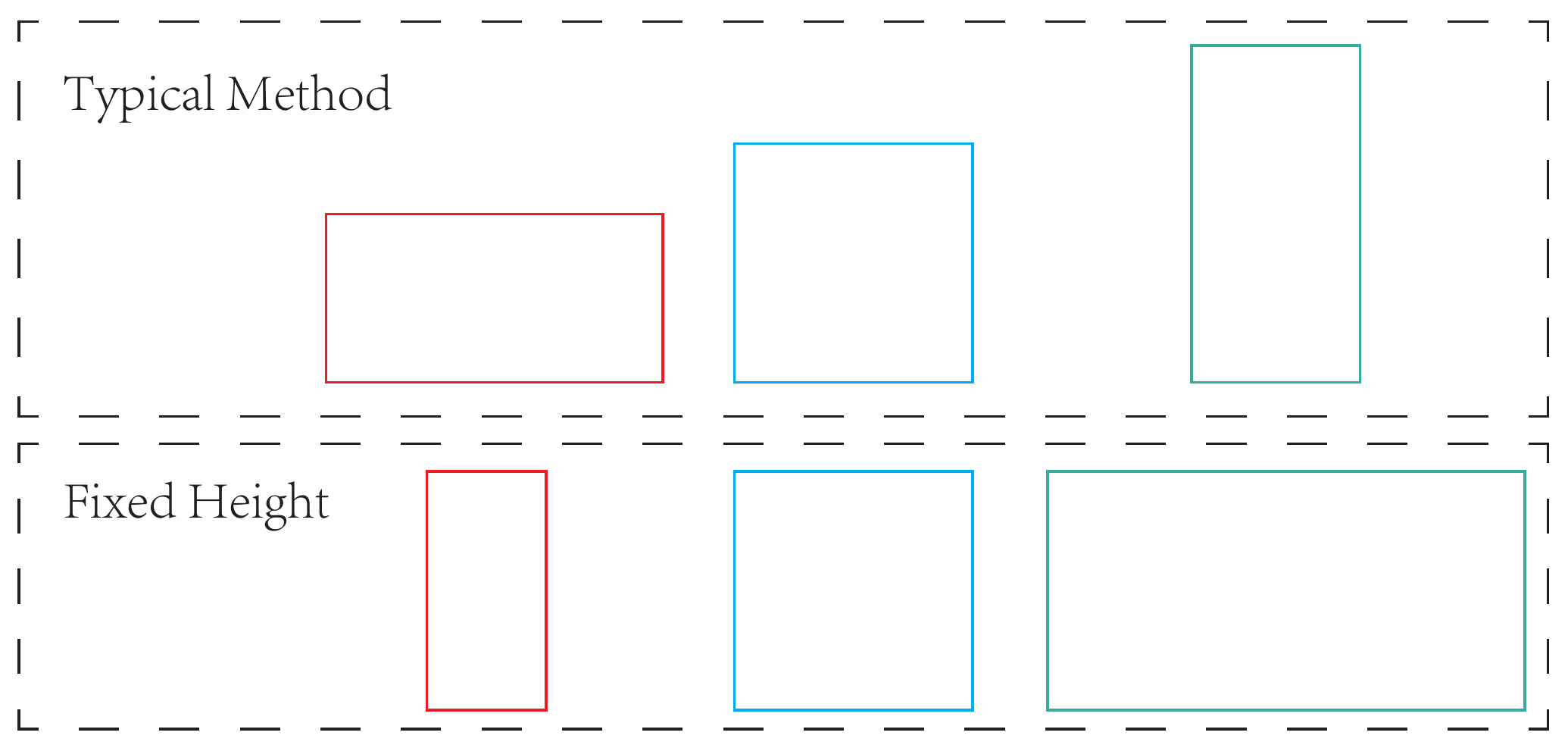}
  \caption{Comparison of typical anchor generation and fixed height anchor generation.}
  \label{fig:anchor_generation}
\end{center}
\end{figure}

In popular two-stage detectors, such as Faster-RCNN and Mask-RCNN, RPN plays a key role in generating region proposals that should be able to cover ground-truth objects as accurate as possible, as shown in Figure~\ref{fig:rcnn_architecture}. Observing that all the columns in a table share the same height and all the rows in a table have an identical width, as shown in Figure~\ref{fig:pubtable1m_sample}, we generate anchors for the columns and rows separately in the step of anchor generation. For the column anchors, we fix the heights for each level of feature pyramids and generate a series of bounding boxes with different aspect ratios, while the generation of row anchors shares a similar process but fixes the width. Notably, anchors for tables and spanning cells can be covered in generating anchors for columns and rows. Figure~\ref{fig:anchor_generation} shows a sample of anchor generation when height is 32, and aspect ratio is [0.5, 1, 2].

\subsection{Cost-sensitive loss}
\label{sec:balancing_loss_hard_example_sampling}
Figure~\ref{fig:rcnn_architecture} shows a generalized architecture of Faster R-CNN, which mainly contains three components, including a Feature Pyramid Network (FPN), a RPN, and a ROIHead. The FPN is used to extract image features usually implemented by a deep CNN network, and the generated features are fed into RPN to generate region proposals. Then the generated features and the region proposals are fed into the ROIHead to fulfill the classification and regression tasks. Our proposed cost-sensitive loss is used in the ROIHead because the generated region proposals are usually imbalanced. Smooth L1 loss $\mathcal{L}_{s}$, which is defined by Equation~\ref{eq:smooth_l1}, is widely used in many two-stage detectors for the regression task. $\mathcal{L}_{s}$ is an extension of L1 loss, smoothing the L1 loss when x is close to 0. However, L1 loss and Smooth L1 loss are sensitive to the size of the bounding box, and it can be beneficial to promote the importance of small objects. Therefore, we further define a cost-sensitive L1 loss $\mathcal{L}_{c}$, which is defined in Equation~\ref{eq:cost_sensitive_l1} and ~\ref{eq:hardness}, in which $\lambda$ is a hyper-parameter used to balance the importance of the number of bounding boxes and the size of bounding boxes, $\alpha$ is hyper-parameter to consider other factors influencing the detection difficulty of each class, which can be set to 0 when other factors influencing the detection difficulty are ignored. $m_i$ and $n_i$ in Equation~\ref{eq:hardness} are the number and size of $i$th class's bounding boxes in a mini-batch during training, and $C$ in Equation~\ref{eq:cost_sensitive_l1} are the total number of classes. At last, we use a weighted sum of each class's smooth L1 loss, in which harder classes are assigned more weights.

\begin{equation}
\label{eq:smooth_l1}
  \mathcal{L}_{s} =
    \begin{cases}
      \frac{x^2}{2\beta} &  if \lvert x \rvert < \beta\\
       \lvert x \rvert - 0.5\beta & otherwise \\
    \end{cases}       
\end{equation}

\begin{equation}
\label{eq:hardness}
l_{i} = (1-\lambda)\frac{m_i}{\sum_{j}{m_j}} + \lambda \frac{n_i}{\sum_{k}{n_k}} + \alpha_i
\end{equation}

\begin{equation}
\label{eq:cost_sensitive_weights}
w_{i} = \frac{exp(-l_i)}{\sum_{j}{exp(-l_j)}}
\end{equation}

\begin{equation}
\label{eq:cost_sensitive_l1}
\mathcal{L}_{c} = \sum_{i=1}^{C} w_i \cdot \mathcal{L}_{s}^{i}
\end{equation}

Similarly, for the classification task in the ROIHead, the most popular loss function is cross-entropy, which is often implemented with a parameter to re-weight the importance of each class. Therefore, we set the weight parameter of cross entropy with the weights as shown in Equation~\ref{eq:cost_sensitive_weights}.

For the TSR problem, as defined in Section~\ref{sec:problem_definition}, there are four types of objects, including table,  table column, table row, and table spanning cell, as shown in Figure~\ref{fig:pubtable1m_sample}. In the defined four types of classes, table spanning cell is usually the hardest type to be detected because spanning cells are the smallest objects and relatively small number of samples. 

\section{Experiments and Analysis}
\label{sec:experiments}

\subsection{Experimental settings and results}
\label{sec:settings_results}
PubTable1M~\cite{smock2021pubtables1m} is a large-scale dataset providing four types of structural units of tables, including table, table column, table row, and table spanning cell, as shown in Figure~\ref{fig:table_sample}. The dataset consists of a training set, validation set, and testing set, each of which contains 758849, 94959, and 93834 samples, respectively.  We use the entire training set for training and the validation set to tune parameters and select the model with the best validation score as the best model.
All the models are implemented with detectron2~\cite{wu2019detectron2}, the MAX\_ITER is set to 180000, and IMS\_PER\_BATCH is set to 16. ResNet50~\cite{he2016deep} and ResNet101~\cite{he2016deep}, both of which are pre-trained with ImageNet~\cite{deng2009imagenet} dataset, are used as the backbone network. Following the popular evaluation metrics in object detection, Average Precision (AP) with different IoU threshold, as defined in Equation~\ref{eq:iou} and Equation~\ref{eq:precision}, are used as the evaluation metric. The experimental results regarding Average Precision  with different IoU threshold are shown in Table~\ref{table:pubtables1m_results}, and Table~\ref{table:pubtables1m_class_results} shows the AP on each class. More specifically, AP in Table~\ref{table:pubtables1m_results} and Table~\ref{table:pubtables1m_class_results} means the average over 10 IoU levels from 0.5 to 0.95 with a step size of 0.05, AP50 and AP75 means the IoU threshold is set as 0.5 and 0.75, respectively. AP$_S$, AP$_M$ and AP$_L$ represent the results on small, medium and large objects respectively, as defined in Equation~\ref{eq:object_scale}. Notably, we do not include the results in study~\cite{smock2021pubtables1m} because of the different experimental settings.

\begin{equation}
\label{eq:iou}
  IoU = \frac{\lvert A \cap B \rvert }{\lvert A \cup B \rvert}
\end{equation}

\begin{equation}
\label{eq:precision}
  Precision = \frac{True Positive}{True Positive + False Positive}
\end{equation}

\begin{equation}
\label{eq:object_scale}
  object =
    \begin{cases}
      small & \text{if area \textgreater $ 32^2$ px}\\
      medium & \text{if $ 32^2$ \textless area \textless $ 64^2$ px} \\
      large &  \text{otherwise} \\
    \end{cases}       
\end{equation}

\begin{figure}[htp]
\begin{center}
  \includegraphics[width=\columnwidth]{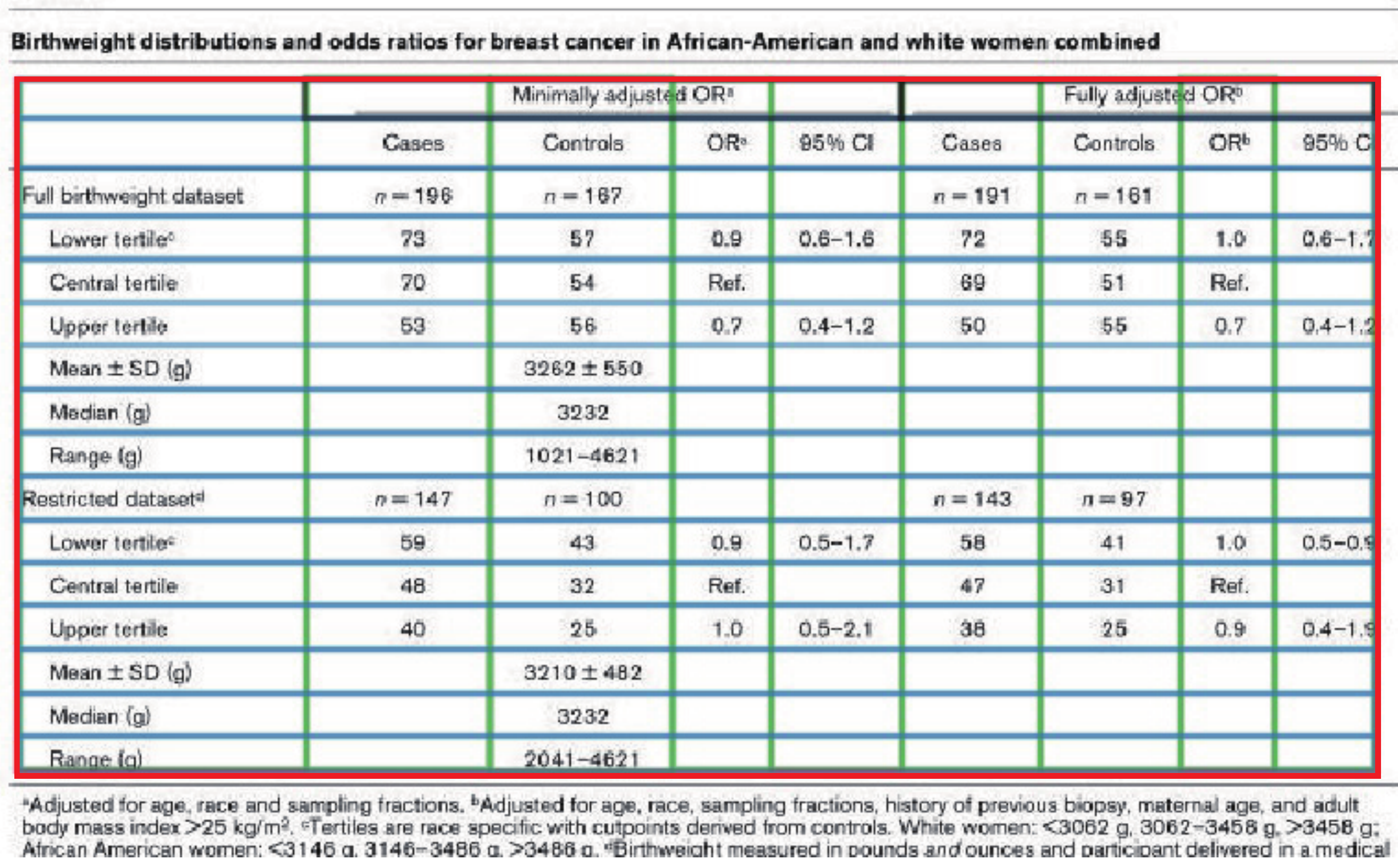}
  \caption{A sample output of the trained Faster-RCNN based model. Notably, red, blue, green, and black bounding boxes denote the table spanning cell, respectively, and the original table does not contain any borderlines.} 
  \label{fig:test_successfull}
\end{center}
\end{figure}

The experimental results show that it is more difficult for detectors to detect table rows and table spanning cells for the original version of Faster-RCNN, but our proposed method can improve the performance of detectors on table rows and table spanning cells consistently, as shown in Table~\ref{table:pubtables1m_class_results}. Meanwhile, Figure~\ref{fig:test_successfull} shows a prediction sample whose input table image does not contain any explicit borderlines, and the trained model can detect all the classes correctly. 

\begin{table}[ht!]
\caption{Experimental results on PubTable1M dataset. $^{\ast}$ and $\dag$ mean that the model is trained with original version of Faster-RCNN and our proposed method, respectively. The batch size is set to 16 for all models in this table.}
\centering
\begin{tabular}{  c | c | c | c | c | c | c }
\hline
\label{table:pubtables1m_results}
Backbone & AP & AP50 & AP75 & AP$_S$ & AP$_M$ & AP$_L$ \\
\hline
ResNet50$^{\ast}$ & $ 89.03 $ & $ 94.72 $ & $ 92.56 $ & $ 71.17 $ & $ 87.89 $ & $ 87.83 $  \\
ResNet101$^{\ast}$ & $ 93.02 $ & $ 96.01 $ & $ 95.03 $ & $ 81.21 $ & $ 92.93 $ & $ 92.26 $  \\
\hline
ResNet50$^\dag$ & $ 90.87 $ & $ 95.80 $ & $ 94.05 $ & $ 70.66 $ & $ 88.82 $ & $ 90.44 $\\
ResNet101$^\dag$ & $ \mathbf{94.79} $ & $ \mathbf{97.46} $ & $ \mathbf{96.42} $ & $ \mathbf{82.55} $ & $ \mathbf{94.20} $ & $ \mathbf{94.53} $  \\
 \hline
\end{tabular}
\end{table}

\begin{table}[ht!]
\caption{Class specific experimental results on PubTable1M datasets. T,  TC, TR, and TSC are short for table, table column, table row, and table spanning cell, respectively. $^{\ast}$ and $\dag$ mean that the model is trained with the original version of Faster-RCNN and our proposed method, respectively. The batch size is set as 16 for all models in this table.}
\centering
\begin{tabular}{  c | c | c | c | c }
\hline
\label{table:pubtables1m_class_results}
Model & T & TC & TR & TSC \\
\hline
ResNet50$^{\ast}$ & $ 98.88 $ & $ 97.32 $ & $ 87.78 $ & $ 73.98 $  \\
ResNet101$^{\ast}$ & $ \mathbf{99.01} $ & $ 98.51 $ & $ 90.99 $ & $ 83.56 $ \\
\hline

ResNet50$^\dag$ & $ 98.89 $ & $ 97.38 $ & $ 93.15 $ & $ 74.05 $ \\
ResNet101$^\dag$ & $ \mathbf{99.01} $ & $ \mathbf{98.53} $ & $ \mathbf{95.81} $ & $ \mathbf{85.81} $ \\
\hline

\end{tabular}
\end{table}

\subsection{Discussion and Analysis}
\label{sec:discussion_analysis}

\subsubsection{The impact of batch size}
Since the proposed cost-sensitive loss is based on the samples in a mini-batch, batch size can influence the performance of the proposed method. In this section, we fixed other parameters and set the batch size to 8, 16, and 32, to analyze the impact of batch size. The experimental results in Table~\ref{table:pubtables1m_batch_size_results} and Table~\ref{table:pubtables1m_batch_size_class_results} show that increasing the number of batch size can be helpful to improve the model performance.
\begin{table}[ht!]
\caption{Experimental results on PubTable1M dataset with different batch size.}
\centering
\begin{tabular}{  c | c | c | c | c | c | c }
\hline
\label{table:pubtables1m_batch_size_results}
Batch Size & AP & AP50 & AP75 & AP$_S$ & AP$_M$ & AP$_L$ \\
\hline
8 & $ 90.00 $ & $ 95.41 $ & $ 93.49 $ & $ 67.92 $ & $ 86.93 $ & $ 89.49 $  \\
16 & $ 90.87 $ & $ 95.80 $ & $ 94.05 $ & $ 70.66 $ & $ 88.82 $ & $ 90.44 $ \\
32 & $ \mathbf{91.51} $ & $ \mathbf{96.30} $ & $ \mathbf{94.61} $ & $ \mathbf{72.92} $ & $ \mathbf{89.44} $ & $ \mathbf{91.03} $  \\
\hline
\end{tabular}
\end{table}

\begin{table}[ht!]
\caption{Class specific experimental results on PubTable1M datasets with different batch sizes. }
\centering
\begin{tabular}{  c | c | c | c | c }
\hline
\label{table:pubtables1m_batch_size_class_results}
Batch Size & T & TC & TR & TSC \\
\hline
8 & $ 98.88 $ & $ 97.04 $ & $ 92.69 $ & $ 71.39 $  \\
16 & $ \mathbf{98.89} $ & $ 97.38 $ & $ 93.15 $ & $ 74.05 $  \\
32 & $ 98.88 $ & $ \mathbf{97.49} $ & $ \mathbf{94.00} $ & $ \mathbf{75.65} $ \\
\hline
\end{tabular}
\end{table}

\subsubsection{The impact of the number of training samples}
As discussed in Section~\ref{sec:experiments}, the PubTab1M dataset contains 758849 images as the training set. However, the images are all from PMCOA corpus, meaning that there is only one data source, and the difference of tables' appearance can be limited. Therefore, we train multiple models using different numbers of training images sampled from the training set randomly to explore whether it is necessary to use such a large training set. The experimental results show that using 1/5 and 1/10 of training samples does not lead to performance degradation significantly for our Faster-RCNN based model. This can be an open issue of how many samples can be enough for different models.

\begin{table}[ht!]
\caption{Experimental results on PubTable1M dataset with different number of training samples.}
\centering
\begin{tabular}{  c | c | c | c | c | c | c }
\hline
\label{table:pubtables1m_sample_size_results}
Samples & AP & AP50 & AP75 & AP$_S$ & AP$_M$ & AP$_L$ \\
\hline
758849 & $ \mathbf{90.87} $ & $ 95.80 $ & $ 94.05 $ & $ \mathbf{70.66} $ & $ \mathbf{88.82} $ & $ \mathbf{90.44} $ \\
151811 & $ 90.79 $ & $ \mathbf{95.95} $ & $ \mathbf{94.17} $ & $ 69.32 $ & $ 87.87 $ & $ 90.27 $ \\
76223 & $ 90.67 $ & $ 95.88 $ & $ 94.09 $ & $ 67.42 $ & $ 87.78 $ & $ 90.17 $  \\
\hline
\end{tabular}
\end{table}

\begin{table}[ht!]
\caption{Class specific experimental results on PubTable1M datasets with different number of training samples. }

\centering
\begin{tabular}{  c | c | c | c | c }
\hline
\label{table:pubtables1m_smaple_size_class_results}
Samples & T & TC & TR & TSC \\
\hline
758849 & $ \mathbf{98.89} $ & $ \mathbf{97.38} $ & $ 93.15 $ & $ \mathbf{74.05} $ \\
151811 & $ 98.88 $ & $ 97.11 $ & $ \mathbf{93.49} $ & $ 73.67 $ \\
76223 & $ 98.88 $ & $ 97.10 $ & $ 93.46 $ & $ 73.25 $ \\
\hline
\end{tabular}
\end{table}

\section{Conclusion and Future Work}
\label{sec:conclusion}

The ICT supply chain is often formulated as a complex Social (Community) Network in the field of supply management, and the relations of participants in the network can be very complex and consists of different interconnections, such as sub-suppliers, suppliers, manufacturers and consumers~\cite{wichmann2016social}. In this study, we focus on the data sharing problem in the complex supply chain network, and propose a solution that can transform unstructured tabular data of components into a structured format to alleviate the problem that the wide usage of tabular data in extremely large volumes exceeds the capacity of human readers. Besides, the proposed method can be helpful in the context of sharing crucial e-component commodities information, because in the ICT supply chain these data are usually summarized and represented within the tables of datasheets. In the proposed solution, we formulate the problem as an object detection problem and build two benchmark models on the PubTab1M dataset. We further define a new low function based on smooth L1 loss considering the number of bounding boxes and the mean size of bounding boxes in every mini-batch during the training process, guiding the detector learn more discriminated features from the hard examples, and propose a new anchor generation method based on the observation that the rows in a table share the same width and the columns in a table share an identical height. The experimental results show that the proposed method can increase the AP of Faster-RCNN models with different backbones by more than 1.5\% consistently. Even though we train and evaluate the model on a large scale, the data source of the dataset is limited. Training and evaluating the model in more domains can be a direction for future work.


\begin{thebibliography}{10}
\providecommand{\url}[1]{#1}
\csname url@samestyle\endcsname
\providecommand{\newblock}{\relax}
\providecommand{\bibinfo}[2]{#2}
\providecommand{\BIBentrySTDinterwordspacing}{\spaceskip=0pt\relax}
\providecommand{\BIBentryALTinterwordstretchfactor}{4}
\providecommand{\BIBentryALTinterwordspacing}{\spaceskip=\fontdimen2\font plus
\BIBentryALTinterwordstretchfactor\fontdimen3\font minus
  \fontdimen4\font\relax}
\providecommand{\BIBforeignlanguage}[2]{{%
\expandafter\ifx\csname l@#1\endcsname\relax
\typeout{** WARNING: IEEEtran.bst: No hyphenation pattern has been}%
\typeout{** loaded for the language `#1'. Using the pattern for}%
\typeout{** the default language instead.}%
\else
\language=\csname l@#1\endcsname
\fi
#2}}
\providecommand{\BIBdecl}{\relax}
\BIBdecl

\bibitem{rodriguez2016social}
R.~Rodriguez-Rodriguez and R.~D. Leon, ``Social network analysis and supply
  chain management,'' \emph{International Journal of Production Management and
  Engineering}, vol.~4, no.~1, pp. 35--40, 2016.

\bibitem{lotfi2013information}
Z.~Lotfi, M.~Mukhtar, S.~Sahran, and A.~T. Zadeh, ``Information sharing in
  supply chain management,'' \emph{Procedia Technology}, vol.~11, pp. 298--304,
  2013.

\bibitem{smock2021pubtables1m}
B.~Smock, R.~Pesala, and R.~Abraham, ``Pub{T}ables-1{M}: Towards comprehensive
  table extraction from unstructured documents,'' \emph{arXiv preprint
  arXiv:2110.00061}, 2021.

\bibitem{easyocr}
JaidedAI, ``Easyocr,'' \url{https://github.com/JaidedAI/EasyOCR.git}, 2022.

\bibitem{du2021pp}
Y.~Du, C.~Li, R.~Guo, C.~Cui, W.~Liu, J.~Zhou, B.~Lu, Y.~Yang, Q.~Liu, X.~Hu
  \emph{et~al.}, ``Pp-ocrv2: Bag of tricks for ultra lightweight ocr system,''
  \emph{arXiv preprint arXiv:2109.03144}, 2021.

\bibitem{ren2015faster}
S.~Ren, K.~He, R.~Girshick, and J.~Sun, ``Faster r-cnn: Towards real-time
  object detection with region proposal networks,'' \emph{Advances in neural
  information processing systems}, vol.~28, 2015.

\bibitem{he2017mask}
K.~He, G.~Gkioxari, P.~Doll{\'a}r, and R.~Girshick, ``Mask r-cnn,'' in
  \emph{Proceedings of the IEEE international conference on computer vision},
  2017, pp. 2961--2969.

\bibitem{prasad2020cascadetabnet}
D.~Prasad, A.~Gadpal, K.~Kapadni, M.~Visave, and K.~Sultanpure,
  ``Cascadetabnet: An approach for end to end table detection and structure
  recognition from image-based documents,'' in \emph{Proceedings of the
  IEEE/CVF conference on computer vision and pattern recognition workshops},
  2020, pp. 572--573.

\bibitem{wang2020deep}
J.~Wang, K.~Sun, T.~Cheng, B.~Jiang, C.~Deng, Y.~Zhao, D.~Liu, Y.~Mu, M.~Tan,
  X.~Wang \emph{et~al.}, ``Deep high-resolution representation learning for
  visual recognition,'' \emph{IEEE transactions on pattern analysis and machine
  intelligence}, vol.~43, no.~10, pp. 3349--3364, 2020.

\bibitem{fernandes2021tabledet}
J.~Fernandes, M.~Simsek, B.~Kantarci, and S.~Khan, ``Tabledet: An end-to-end
  deep learning approach for table detection and table image classification in
  data sheet images,'' \emph{Neurocomputing}, 2021.

\bibitem{schreiber2017deepdesrt}
S.~Schreiber, S.~Agne, I.~Wolf, A.~Dengel, and S.~Ahmed, ``Deepdesrt: Deep
  learning for detection and structure recognition of tables in document
  images,'' in \emph{Intl. Conf. on document analysis and recognition},
  vol.~1.\hskip 1em plus 0.5em minus 0.4em\relax IEEE, 2017, pp. 1162--1167.

\bibitem{xiao2022table}
B.~Xiao, M.~Simsek, B.~Kantarci, and A.~A. Alkheir, ``Table structure
  recognition with conditional attention,'' \emph{arXiv preprint
  arXiv:2203.03819}, 2022.

\bibitem{chi2019complicated}
Z.~Chi, H.~Huang, H.-D. Xu, H.~Yu, W.~Yin, and X.-L. Mao, ``Complicated table
  structure recognition,'' \emph{arXiv preprint arXiv:1908.04729}, 2019.

\bibitem{khan2017cost}
S.~H. Khan, M.~Hayat, M.~Bennamoun, F.~A. Sohel, and R.~Togneri,
  ``Cost-sensitive learning of deep feature representations from imbalanced
  data,'' \emph{IEEE transactions on neural networks and learning systems},
  vol.~29, no.~8, pp. 3573--3587, 2017.

\bibitem{shrivastava2016training}
A.~Shrivastava, A.~Gupta, and R.~Girshick, ``Training region-based object
  detectors with online hard example mining,'' in \emph{Proceedings of the IEEE
  conference on computer vision and pattern recognition}, 2016, pp. 761--769.

\bibitem{pang2019libra}
J.~Pang, K.~Chen, J.~Shi, H.~Feng, W.~Ouyang, and D.~Lin, ``Libra r-cnn:
  Towards balanced learning for object detection,'' in \emph{Proceedings of the
  IEEE/CVF conference on computer vision and pattern recognition}, 2019, pp.
  821--830.

\bibitem{oksuz2020imbalance}
K.~Oksuz, B.~C. Cam, S.~Kalkan, and E.~Akbas, ``Imbalance problems in object
  detection: A review,'' \emph{IEEE transactions on pattern analysis and
  machine intelligence}, vol.~43, no.~10, pp. 3388--3415, 2020.

\bibitem{wu2019detectron2}
Y.~Wu, A.~Kirillov, F.~Massa, W.-Y. Lo, and R.~Girshick, ``Detectron2,''
  \url{https://github.com/facebookresearch/detectron2}, 2019.

\bibitem{he2016deep}
K.~He, X.~Zhang, S.~Ren, and J.~Sun, ``Deep residual learning for image
  recognition,'' in \emph{Proceedings of the IEEE conference on computer vision
  and pattern recognition}, 2016, pp. 770--778.

\bibitem{deng2009imagenet}
J.~Deng, W.~Dong, R.~Socher, L.-J. Li, K.~Li, and L.~Fei-Fei, ``Imagenet: A
  large-scale hierarchical image database,'' in \emph{2009 IEEE conference on
  computer vision and pattern recognition}.\hskip 1em plus 0.5em minus
  0.4em\relax Ieee, 2009, pp. 248--255.

\bibitem{wichmann2016social}
B.~K. Wichmann and L.~Kaufmann, ``Social network analysis in supply chain
  management research,'' \emph{International Journal of Physical Distribution
  \& Logistics Management}, 2016.

\end{thebibliography}
\bibliographystyle{IEEEtran}

\end{document}